\documentclass[10pt,twocolumn,letterpaper]{article}

\usepackage{cvpr}
\usepackage{times}
\usepackage{epsfig}
\usepackage{graphicx}
\usepackage{subcaption}
\usepackage{multirow}
\graphicspath{{./figures/}}
\usepackage{enumitem}
\usepackage{amsmath}
\usepackage{amssymb}
\usepackage{amsthm} 
\usepackage{resizegather}
\usepackage{mathtools}
\usepackage{ctable} 
\DeclarePairedDelimiter\ceil{\lceil}{\rceil}
\DeclarePairedDelimiter\floor{\lfloor}{\rfloor}

\DeclarePairedDelimiter\abs{\lvert}{\rvert}

\newcommand{\resize}[2]{Bilinear_{{#1} \rightarrow {#2}}}
\newcommand{\frackernel}[1]{k_{{\scriptstyle AFDC}}^{#1}}
\newcommand{\kernel}[3]{k_{{\scriptstyle ({#1},{#2})}}^{{#3}}}
\newcommand{\conv}{*}
\newcommand{\ratio}{\frac{w}{h}}

\newcommand{\weight}[2]{w_{({#1}, {#2})}}

\newcommand{\featurev}[1]{\mathbf{F}_{#1}}

\newtheorem{theorem}{Theorem}
\newtheorem{lemma}{Lemma}
\newtheorem*{definition}{Definition}

\usepackage[breaklinks=true,bookmarks=false]{hyperref}
\usepackage[capitalise]{cleveref}

\cvprfinalcopy 

\def\cvprPaperID{8932} 

\ifcvprfinal\fi
\begin{document}

\title{Adaptive Fractional Dilated Convolution Network \\for Image Aesthetics Assessment}

\author{Qiuyu Chen$^1$, Wei Zhang$^{2}$, Ning Zhou$^3$, Peng Lei$^3$, Yi Xu$^2$, Yu Zheng$^4$, Jianping Fan$^1$\\
    {\small $^1$Department of Computer Science, UNC Charlotte}\\
    {\small $^2$Shanghai Key Laboratory of Intelligent Information Processing, School of Computer Science, Fudan University}\\
    {\small $^3$Amazon Lab126} \\
    {\small $^4$School of Cyber Engineering, Xidian University} \\
    {\tt\small \{qchen12,jfan\}@uncc.edu, \{weizh,yxu17\}@fudan.edu.cn}\\
    {\tt\small \{ningzho,leipeng\}@amazon.com, yuzheng.xidian@gmail.com}
}

\maketitle
\thispagestyle{empty}

\begin{abstract} 
    To leverage deep learning for image aesthetics assessment, one critical but unsolved issue is how to seamlessly incorporate the information of image aspect ratios to learn more robust models.
    In this paper, an adaptive fractional dilated convolution (AFDC), which is aspect-ratio-embedded, composition-preserving and parameter-free, is developed to tackle this issue natively in convolutional kernel level.
    Specifically, the fractional dilated kernel is adaptively constructed according to the image aspect ratios, where the interpolation of nearest two integer dilated kernels are used to cope with the misalignment of fractional sampling.
    Moreover, we provide a concise formulation for mini-batch training and utilize a grouping strategy to reduce computational overhead.
    As a result, it can be easily implemented by common deep learning libraries and plugged into popular CNN architectures in a computation-efficient manner.
    Our experimental results demonstrate that our proposed method achieves state-of-the-art performance on image aesthetics assessment over the AVA dataset~\cite{murray-ava}.
\end{abstract}

\section{Introduction}\label{sec:introduction}

This paper addresses image aesthetics assessment where the goal is to predict the given image an aesthetic score. 
Automatic image aesthetics assessment has many applications such as album photo recommendation, auxiliary photo editing, and multi-shot photo selection. 
The task is challenging because it entails computations of both global cues (\eg scene, exposure control, color combination, etc) and localization information (composition, photographic angle, etc).

Early approaches extract aesthetic features according to photographic rules (lighting, contrast) and global image composition (symmetry, rule of thirds), which require extensive manual designs~\cite{traditional01,traditional06,traditional02,traditional05,tradistional04,traditional03}.
However, manual design for such aesthetic features is not a trivial task even for experienced photographers.
Recent work adopts deep convolutional neural networks for image aesthetics assessment by learning models in an end-to-end fashion.
The models mainly use three types of formulations: binary classification labels~\cite{kao2016hierarchical, ma-multipath,mai-adaptive,wang-multiscene,ren-peronalized,zeng-unified,sheng-attention}, scores~\cite{murray-huberloss,talebi-nima,hosu-multi}, and rankings~\cite{kong-ranking,schwarz-tripleloss}.

\begin{figure}[t]
        \centering
        \includegraphics[width=.9\linewidth]{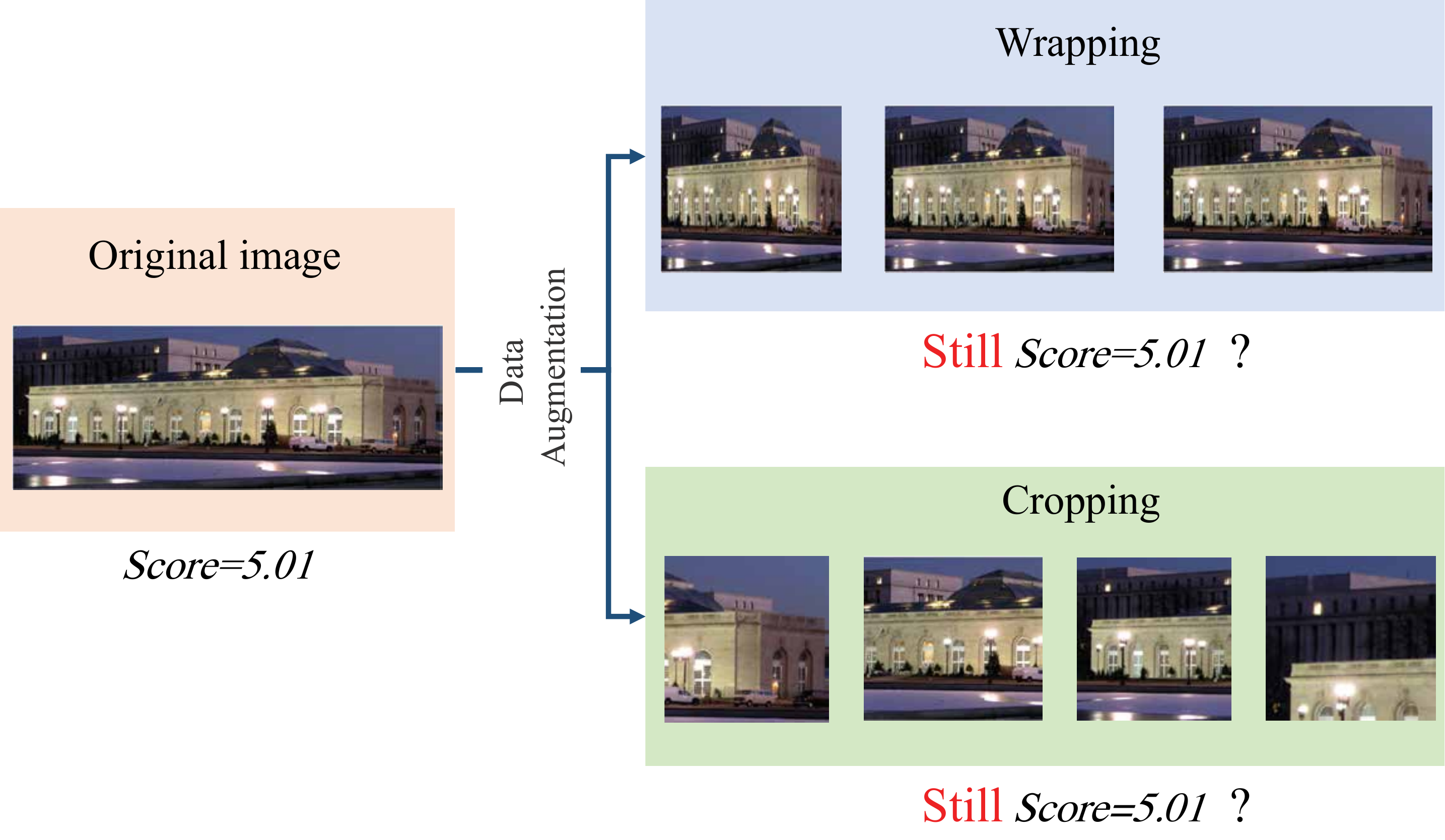}
        \vspace*{-0.30cm}
        \caption{\bf Image warping and cropping are widely used for data augmentation, but they alter the object aspect ratios and composition, causing different aesthetics perceptions. Assigning the groundtruth aesthetic score of the original image to the altered image may introduce label noise and deteriorate the discriminative ability.}
        \label{fig:label-noise}
        \vspace*{-0.70cm}
\end{figure}

\begin{figure*}[t]
    \begin{center}
        \includegraphics[width=\linewidth]{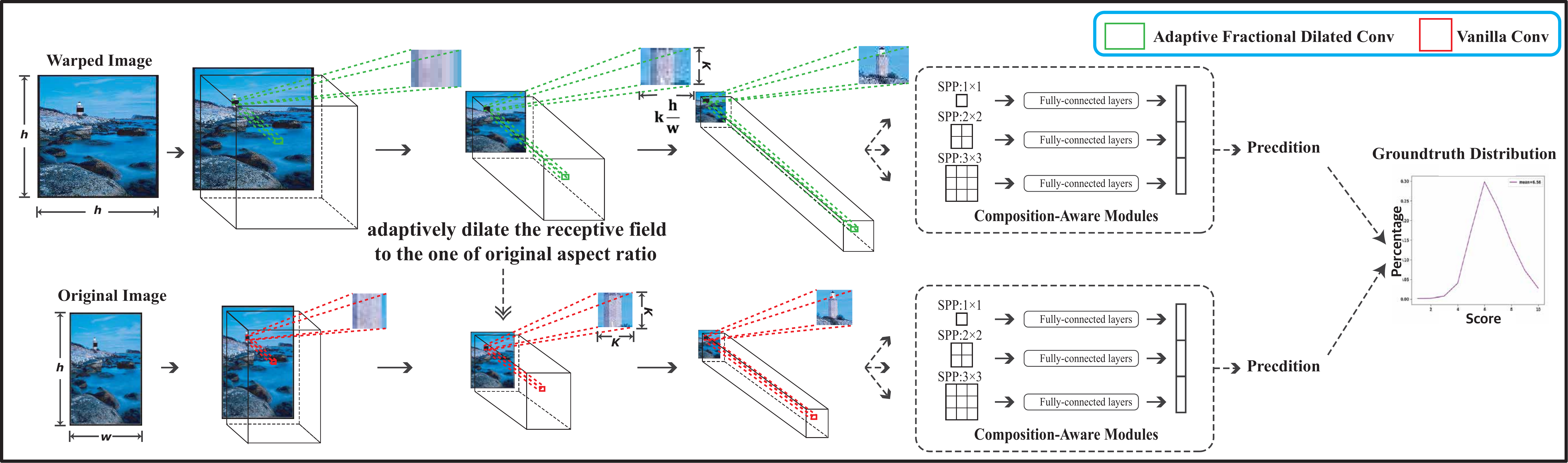}
    \end{center}
    \vspace*{-0.40cm}
    \caption{\bf Overview of adaptive fractional dilated CNN (above) and the comparison with vanilla CNN (below): Each fractional dilated Conv (above) operated on wrapped input adaptively dilates the same receptive field as the vanilla Conv (below) operated on the original image. It thus helps with the problems: (a) Becomes mini-batch compatible by composition-preserving warping instead of feeding original-size image (b) Preserves aesthetic features related to aspect ratios by adaptive kernel dilation.}
    \label{fig:overview}
    \vspace*{-0.55cm}
\end{figure*}

In the aforementioned methods, the backbone networks are usually adopted from an image classification network.
The data augmentation methods, \ie image cropping and warping, are widely used for preventing overfitting in the image recognition task.
However, a shortcoming is that the compositions and object aspect ratios are altered, which may introduce label noise and harm the task of aesthetics assessment (\cref{fig:label-noise}).
A succinct solution proposed in MNA-CNN~\cite{mai-adaptive} is to feed one original-size image into the network at a time during training and test (bottom stream in \cref{fig:overview}). A major constraint of the approach is that images with different aspect ratios cannot be concatenated into batches because the aspect ratio of each image should be preserved. Thus it slows down the training and inference.

In this paper, we aim to develop a novel adaptive fractional dilated convolution that is mini-batch compatible. 
As shown in the top row in \cref{fig:overview}, our network adaptively dilates the convolution kernels to the composition-preserving warped images according to the image aspect ratios
such that the effective receipt field of each dilated convolution kernel is the same as the regular one.
Specifically, as illustrated in \cref{fig:frac-dilated-conv-kernel}, the fractional dilated convolution kernel is adaptively interpolated by the nearest two integer dilated kernels with the same kernel parameters.
Thus no extra learning parameters are introduced. 

The {\bf benefits} of our method can be summarized as follows: 
(a) By embedding the information of aspect ratios to construct the convolution layers adaptively, it can explicitly relate the aesthetic perception to the image aspect ratios while preserving the composition;
(b) It is parameter-free and thus can be easily plugged into the popular network architectures;
(c) Through the deduction, we show that our proposed method can be mini-batch compatible and easily implemented by common deep learning libraries (\eg PyTorch, Tensorflow);
(d) A grouping strategy is introduced to reduce the computational overhead for efficient training/inference;
(e) We achieve state-of-the-art performance for image aesthetics assessment on the AVA dataset~\cite{murray-ava}. 

\section{Related Work}

In this section, we provide a brief review of some of the most relevant works on: (a) image aesthetics assessment; (b) preserving image aspect ratios and compositions; (c) dilated convolution; (d) dynamic kernels.

\noindent{\bf Image Aesthetics Assessment.} The existing methods on image aesthetics assessment can be mainly categorized into three formulations: (1) \textit{Binary (or mean) aesthetic label}:
Kao \etal~\cite{kao2016hierarchical} propose a multi-task CNN, A\&C CNN, which jointly learns both the category classification and the aesthetic perception.
Mai \etal~\cite{mai-adaptive} address the composition problem in image aesthetics assessment and aggregates multiple sub-networks with different sizes of adaptive pooling layer.
Ma \etal~\cite{ma-multipath} feed the patches sampled from the saliency map of the original image into VGG16~\cite{simonyan-vgg} with an aggregation layer, where a layer-aware subnet considering path localizations is leveraged to get the final prediction.
Sheng \etal~\cite{sheng-attention} assign adaptively larger weights to meaningful training cropping patches according to the prediction errors during the training and aggregate the multi-patch predictions during the test.
Hosu \etal~\cite{hosu-multi} propose to incorporate multi-level spatially polled features from the intermediate blocks in a computation efficient manner.
(2) \textit{Ranking score}:
Instead of classification or regression formulations, a joint loss of Euclidean and ranking~\cite{kong-ranking} is proposed and a triplet ranking loss~\cite{schwarz-tripleloss} is developed.
(3) \textit{Score distribution}:
To address the ordered score distribution, Hossein Talebi and Peyman Milanfar~\cite{talebi-nima} introduce Earth Mover's Distance as a loss function to train 10-scale score distribution.
Since the image aesthetics is a subjective property and outlier opinions may appear, Naila Murray and Albert Gordo~\cite{murray-huberloss} introduce Huber Loss to train 10-scale score distribution.
Besides using the mean score of multiple raters, Ren \etal~\cite{ren-peronalized} propose a sub-network to learn a personal rating offset along with the generic aesthetic network and output the personalized score prediction.

\noindent{\bf Preserving Image Aspect Ratios and Compositions.}
Multi-patch sampling over the original images is used to preserve the aspect ratios and proves to be effective~\cite{ma-multipath,sheng-attention,hosu-multi}.
A major concern is that sampling patches from the original image may alter essential aesthetic factors (color histogram, object-background ratio) of the original image and the complete aesthetics features are lost.
In contrast, our proposed method adaptively restores the original receptive fields from the composition-preserving warping images in an end-to-end fashion.
The approach of MNA-CNN~\cite{mai-adaptive} is the most related to ours, as they proposed to preserve image aspect ratios and compositions by feeding the original image into the network, one at a time.
A major constraint of the approach is that images with different aspect ratios cannot be concatenated into batches because the aspect ratio of each image should be preserved.
Thus it tends to slow down the training and inference processes.
On the other hand, our proposed method is mini-batch compatible and can be easily implemented by common deep learning libraries. 

\noindent{\bf Dilated Convolution.}
Our adaptive fractional dilated convolution is motivated by the 
dilated convolution~\cite{yu-dilatedconv} and atrous convolution~\cite{chen-deeplab} in semantic segmentation, 
but it differs from them in several aspects:
(1) Our adaptive fractional dilated convolution is to restore the receptive fields for warped images to the same as regular convolution for original images, while dilated convolution is proposed to retrain the large receptive without down-sampling.
(2) The dilation rate can be fractional in our method.
(3) The construction of fractional dilated kernel is dynamic respecting the aspect ratios.

\noindent{\bf Dynamic Kernels.}
Deformable convolution~\cite{dai-deformable} is proposed to construct the receptive fields dynamically and adaptively by learning better sampling in the convolutional layer.
Our proposed method differs from deformable convolution in two folds:
(a) The deformable convolution is proposed to learn better sampling in the convolutional layers, whereas our method adapts the receptive fields into the original aspect ratios. Therefore, our proposed method is parameter-free while the deformable convolution requires parameterized layers to predict the sampling indices. 
(b) Our method provides a concise formula for mini-batch training and it can be easily implemented by the common deep learning frameworks. On the other hand, the deformable convolution needs to rewrite the convolution operation in CUDA and tends to be slow due to the indexing operation.

\section{Adaptive Fractional Dilated Convolution}
In this section, we first introduce the adaptive kernel interpolation to tackle the misalignment due to fractional sampling in the proposed method.
We then derive a concise formulation for it in the setting of mini-batch and discuss their computational overhead.
Finally, we describe the loss function and an additional composition-aware structure for the composition-preserving warping batch.

\subsection{Adaptive Kernel Interpolation}\label{sec:FracDilated-conv}

As stated in \cref{sec:introduction}, cropping modifies the composition of the original image and causes the loss of some critical aesthetics information.
As a result,  image cropping introduces somewhat label noises in the training stage.
To preserve the composition, we firstly warp the image into a fixed size.
For network training, such a simple image warping approach suffers from the problem of overfitting due to the absence of data augmentation.
Motivated by SPP~\cite{he-spp}, we adopt random-size warping during the training stage and feed the mini-batch into the networks with global pooling or SPP modules, which can naturally handle arbitrary-size batch inputs.
Overall, the random-size warping provides effective data augmentation for training scale-invariant networks while preserving the image compositions.

To cope with the distortion induced by warping, the receptive field of the convolution kernel should be consistent with the receptive field of the convolution kernel that is operated on the image with original aspect ratio.
Our proposed approach tackles the distortion issue by adaptively dilating the kernels to the original aspect ratio, as illustrated in \cref{fig:overview}.
Since the aspect ratio could be fractional, the dilation rate could be a fraction as well.
To tackle the misalignment of feature sampling, we use the linear interpolation of two nearest integer dilation rates to construct the fractional dilation kernel.

\begin{figure}
    \centering
    \includegraphics[width=\linewidth]{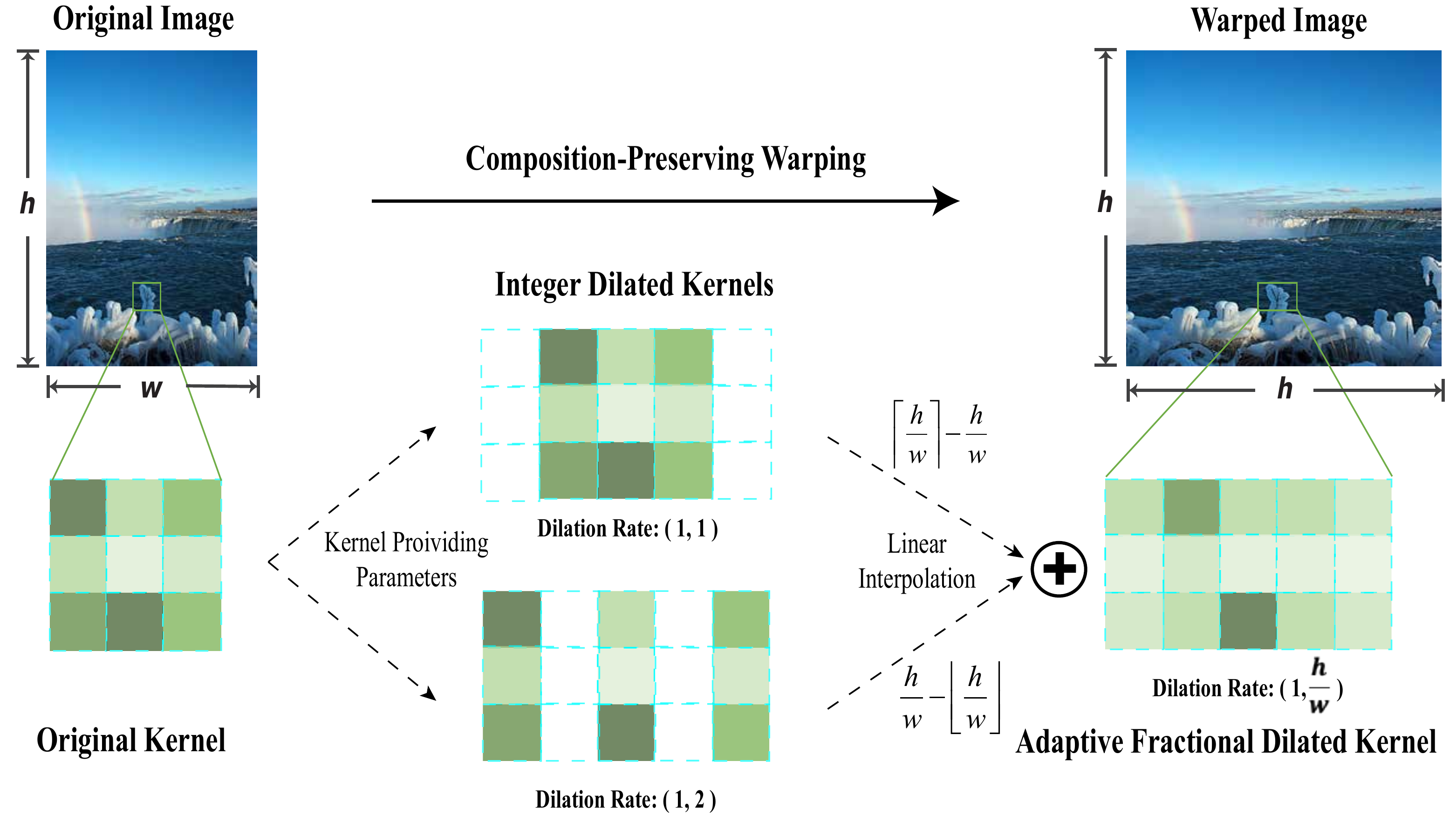}
    \vspace*{-0.45cm}
    \caption{\bf Illustration of kernel interpolation: linear interpolation of the nearest two integer dilated kernels shared same kernel parameters are used to tackle the sampling misalignment from fractional dilation rates.}
    \label{fig:frac-dilated-conv-kernel} 
    \vspace*{-0.55cm}
\end{figure}

Suppose that $w$ and $h$ represent the width and height of original images, respectively.
If $h>w$ and $\frac{h}{w}$ is not a integer, as illustrated in \cref{fig:frac-dilated-conv-kernel}, AFDC (adaptive fractional dilated convolution) kernel $\frackernel{n}$ in $n$-th layer is constructed as:
\begin{equation}\label{eq:frac-dilated-kernel}
\frackernel{n} = {\scriptstyle(\ceil*{r}-r)}
\kernel{1}{\floor*{r}}{n}
+
{\scriptstyle(r-\floor*{r})}
\kernel{1}{\ceil*{r}}{n}
\end{equation}
where  $r=\frac{h}{w}$.
For any non-integer $r$, it is in the interval $\big[\floor*{r},\ceil*{r}\big]$ whose length is equal to 1.
$\floor*{r}$ and $\ceil*{r}$ are two integers nearest to $r$. 
$\kernel{1}{\floor*{r}}{n}$ and $\kernel{1}{\ceil*{r}}{n}$ are two dilated kernels with the nearest integer dilation rates $\floor*{r}$ and $\ceil*{r}$ for $n$th layer, respectively.
More specifically, as shown in \cref{fig:frac-dilated-conv-kernel}, $r\in [1,2]$, $\floor*{r}=1$, $\ceil*{r}=2$.
We note that both $\kernel{1}{1}{n}$ and $\kernel{1}{2}{n}$ inherit the same learning parameters from the original kernel.

Likewise, if $w>h$ and $\frac{w}{h}$ is not an integer, then we choose:
\begin{equation}\label{eq:frac-dilated-kernel_2}
\frackernel{n} = {\scriptstyle(\ceil*{r}-r)}
\kernel{\floor*{r}}{1}{n}
+
{\scriptstyle(r-\floor*{r})}
\kernel{\ceil*{r}}{1}{n}
\end{equation}

If $r=\frac{h}{w}$ is an integer, it is enough for us to employ integer dilated kernel.

Therefore, the fractional dilated kernel is adaptively constructed for each image with respect to $w$ and $h$ as shown in \cref{fig:frac-dilated-conv-kernel}.
In addition, all the integer dilation kernels share the same kernel parameters and thus no extra learning parameters are introduced.

\subsection{Mini-Batch Computation and Implementation}\label{sec:mini-batch}

\begin{figure}
    \centering
    \includegraphics[width=.8\linewidth]{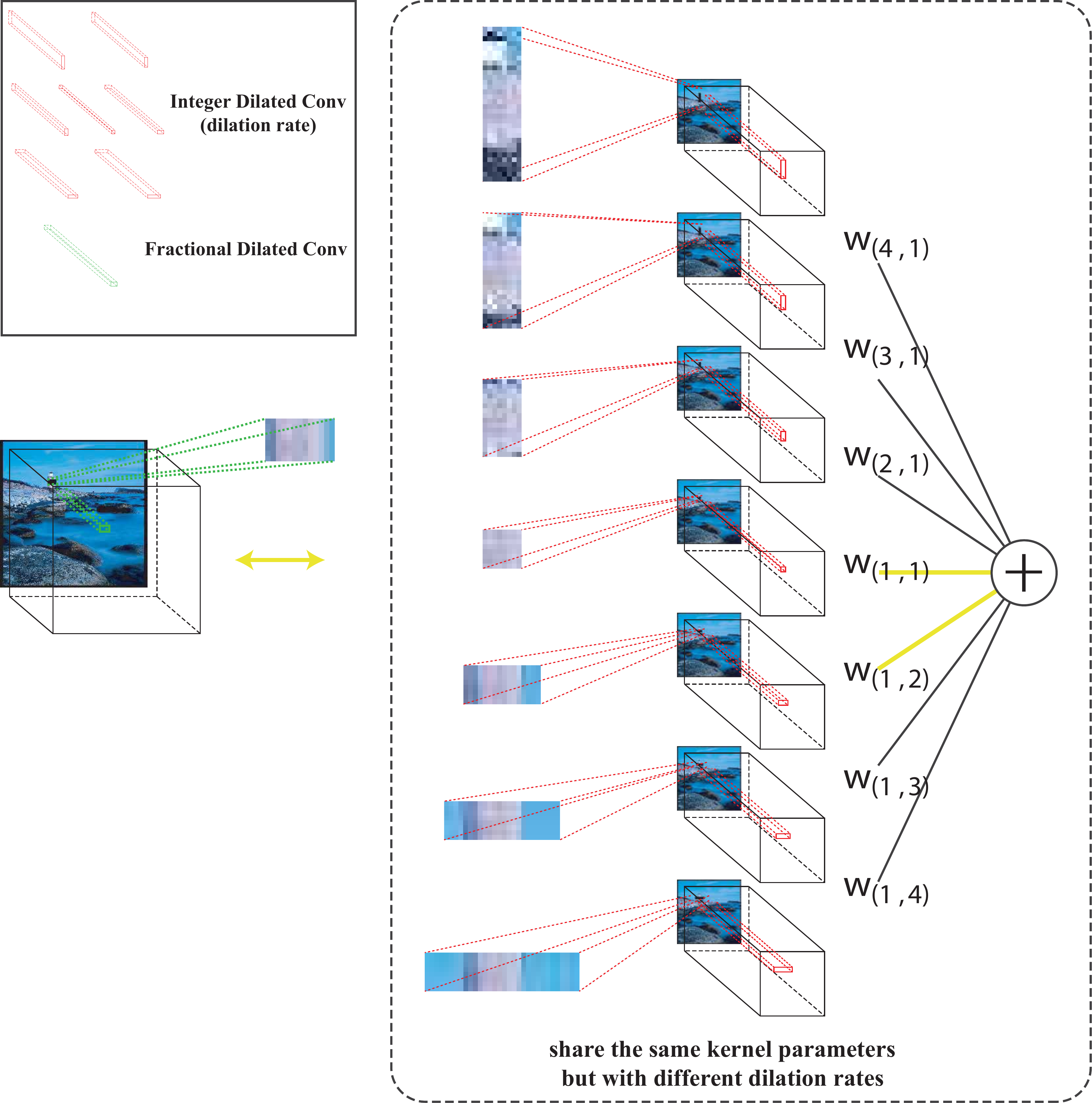}
    \caption{\bf Illustration for mini-batch compatibility: the distributive property of convolution operation (\cf \cref{eq:frac-dilated-conv}) makes the fractional dilated conv easily implemented and compatible for mini-batch computation with a zero-padded weight vector/matrix (\cf \cref{eq:weight-vector})}
    \vspace*{-0.45cm}
    \label{fig:zero-padding}
    \vspace*{-0.20cm}
\end{figure}

To implement the dynamic kernel interpolation in \cref{eq:frac-dilated-kernel} and \cref{eq:frac-dilated-kernel_2} directly, we need to rewrite the kernel-level code due to the diverse kernels in mini-batch.
However, through the following deduction, we show that the proposed method can be easily implemented by common deep learning libraries, \eg PyTorch and TensorFlow.

Using the distributive property of convolution operation, the transformation of the feature maps generated by the adaptive fractional dilated Conv kernels in \cref{eq:frac-dilated-kernel} can be formulated as:
\begin{equation}\label{eq:frac-dilated-conv}
\resizebox{.9\linewidth}{!}
{
    $
    \begin{split}
    f_{n+1}&=\frackernel{n} \conv f_n \\
    &=\bigg[
    {\scriptstyle(\ceil*{\ratio}-\ratio)}
    \kernel{1}{\floor*{\ratio}}{n}
    +
    {\scriptstyle(\ratio-\floor*{\ratio})}
    \kernel{1}{\ceil*{\ratio}}{n}
    \bigg] \conv f_n \\
    &={\scriptstyle(\ceil*{\ratio}-\ratio)}
    \kernel{1}{\floor*{\ratio}}{n}
    \conv f_n +
    {\scriptstyle(\ratio-\floor*{\ratio})}
    \kernel{1}{\ceil*{\ratio}}{n}
    \conv f_n
    \end{split}
    $
}
\end{equation}
where $f_{n}$ denotes the feature maps for the $n$th layer and $\conv$ denotes convolution.

In mini-batch training and inference, we can construct multiple kernels with different dilation rates $(rate_k^i, rate_k^j)$ from the same kernel parameters and then use a zero-padded interpolation weight vector $\mathbf{w}$ to compute the operation adaptively for each image as:
\begin{equation}\label{eq:frac-dilated-conv-weightv}
\begin{split}
f_{n+1}&=\frackernel{n}
\conv f_{n} \\
&= \sum_{k} \weight{rate_k^i}{rate_k^j}
\kernel{rate_k^i}{rate_k^j}{n}
\conv f_{n}\\
&=\mathbf{w}\widetilde{\mathbf{f}}_{n}
\end{split}
\end{equation}
which is just the inner product of two vectors:
\begin{equation}\label{eq:weight-vector}
\mathbf{w}=[\weight{rate_1^i}{rate_1^j},...,\weight{rate_K^i}{rate_K^j}]
\end{equation}
and
\begin{equation}
\widetilde{\mathbf{f}}_{n}=[\kernel{rate_1^i}{rate_1^j}{n}
\conv f_{n},...,\kernel{rate_K^i}{rate_K^j}{n}
\conv f_{n}]^{\top}
\end{equation}
where the number of dilation kernels is $K$. As shown in \cref{fig:zero-padding}, the interpolation weight $\weight{rate_k^i}{rate_k^j}$ for each instance is either $\weight{rate_k^i}{1}$ or $\weight{1}{rate_k^j}$,  defined as follows:
\begin{equation}\label{eq:weights}
\begin{split}
\weight{rate^i}{1} &=
\begin{cases}
r - (rate^i - 1), &\text{if } rate^i - r \in [0, 1) \\
(rate^i + 1) - r, &\text{if } rate^i - r \in (-1, 0) \\
0, &\text{else}
\end{cases} \\
\weight{1}{rate^j}&=
\begin{cases}
r - (rate^j - 1), &\text{if } rate^j-r \in [0, 1)\\
(rate^j + 1) - r, &\text{if } rate^j-r \in (-1, 0) \\
0, &\text{else}
\end{cases}
\end{split}
\end{equation}
In mini-batch, suppose that batch size is $B$, then the  $n+1$th feature maps $\featurev{n+1}$ can be formulated as:
\begin{equation}\label{eq:frac-dilated-conv-batch}
\featurev{n+1} =[\mathbf{f}^1_{n+1},...,\mathbf{f}^B_{n+1}]=[\mathbf{w}^1\widetilde{\mathbf{f}}^1_{n},...,\mathbf{w}^B\widetilde{\mathbf{f}}^B_{n}]
\end{equation}
The computation of the above $[\widetilde{\mathbf{f}}^1_{n},...,\widetilde{\mathbf{f}}^B_{n}]$ can be done efficiently in the mini-batch as:
\begin{equation}\label{eq:dilation-group-maps-batch}
\resizebox{.9\linewidth}{!}
{$
    \begin{bmatrix}
    \kernel{rate_1^i}{rate_1^j}{n} \conv \featurev{n}, &
    \kernel{rate_2^i}{rate_2^j}{n} \conv \featurev{n}, &
    \dots, &
    \kernel{rate_K^i}{rate_K^j}{n} \conv \featurev{n}
    \end{bmatrix}^{\top}
    $}
\end{equation}

We note that the activation function and batch normalization are omitted in the formulas for concise illustration. 

The formula in \cref{eq:frac-dilated-conv-batch} can be interpreted as a dot production followed by a sum reduction between interpolation weight matrix $\mathbf{W}$ and \cref{eq:dilation-group-maps-batch}, which thus can be efficiently implemented by common deep learning frameworks (Pytorch, Tensorflow, etc.).
Each integer dilated Conv, $\kernel{rate_k^i}{rate_k^j}{n} \conv \featurev{n}$ in \cref{eq:dilation-group-maps-batch}, is computed as a normal dilated Conv layer with the shared learning parameters.

\begin{figure}
    \centering
    \includegraphics[width=.8\linewidth]{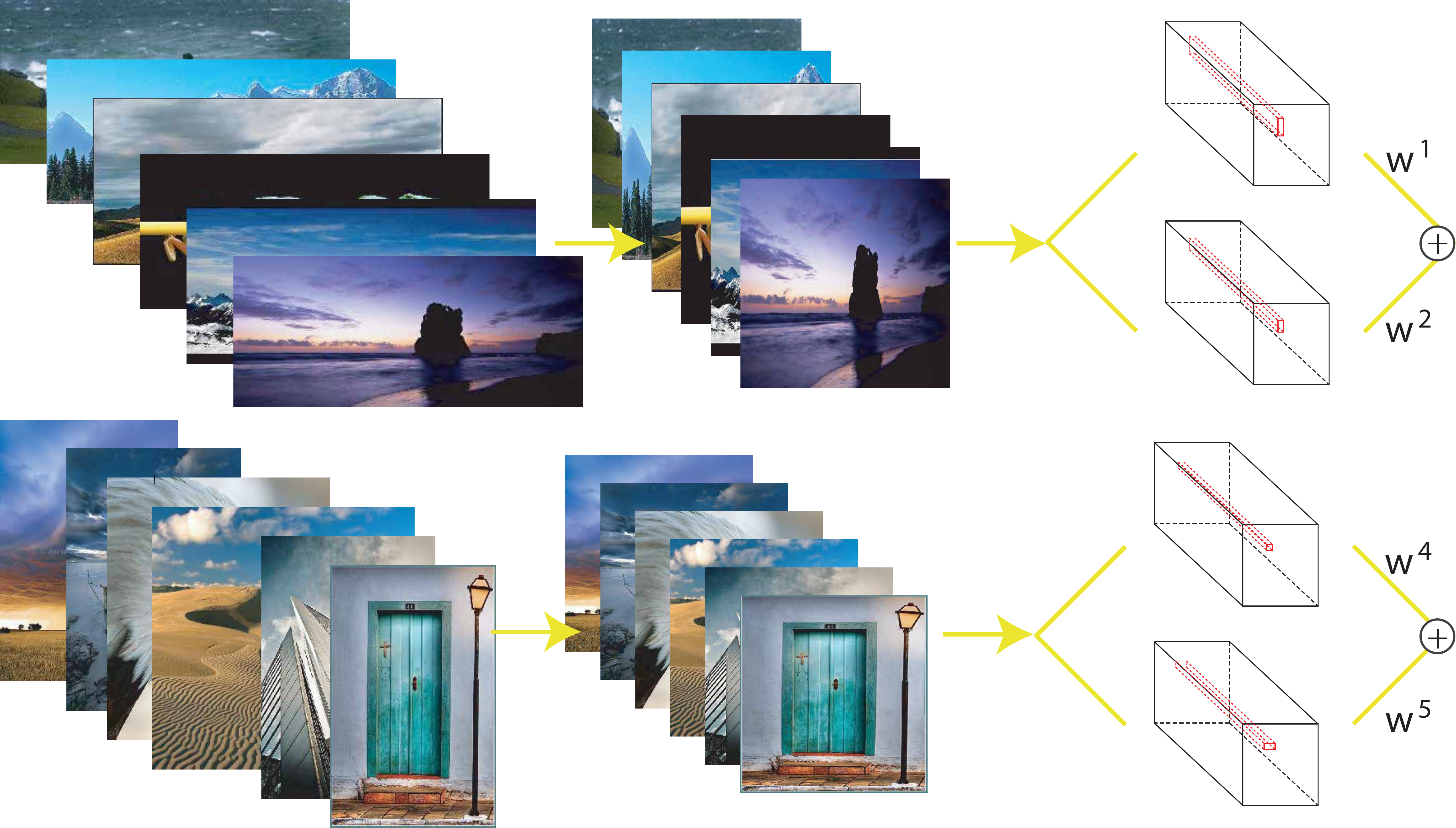}
    \vspace*{-0.30cm}
    \caption{ \bf Grouping strategy to reduce computational overhead: The integer dilated Convs can be shared by properly grouped images according to aspect ratios.}
    \label{fig:grouping-strategy}
    \vspace*{-0.35cm}
\end{figure}

\begin{table}
    \begin{center}
        \scalebox{0.8}{
        \begin{tabular}{c|c|c|c|c}
            \toprule
            Network & \#Params & \#Mult-Adds & Speed (train) & Speed (test) \\
            \specialrule{1.5pt}{1pt}{1pt}
            VGG16 & 138M & 15.3G & 8.14 it/s & 12.91 it/s \\
            2-dilation & 138M & 30.7G & 2.70 it/s & 3.85 it/s \\
            7-dilation & 138M & 109.1G & 0.73 it/s & 0.93 it/s \\
            \midrule
            ResNet50 & 25.6M & 3.5G & 12.49 it/s & 22.80 it/s \\
            2-dilation & 25.6M & 5.6G & 8.32 it/s &  14.81 it/s \\
            2-dilation* & 25.6M & 6.5G & 6.20 it/s & 9.88 it/s \\
            7-dilation & 25.6M & 10.6G & 3.22 it/s & 5.28 it/s \\
            7-dilation* & 25.6M & 18.8G & 2.08 it/s & 3.12 it/s \\
            \bottomrule
        \end{tabular}
        }
        \vspace*{-0.25cm}
        \caption{\bf Computation comparison: training batch size is set to 16, test batch size is set to 32. The speed is the average result for 100 iterations from the test on single GTX 1080Ti. The fractional dilated Conv is embedded for all BottleNets in ResNet50 while * denotes additional embedding dilation for the first $7\times7$ Conv layer as well.}
        \label{tab:comp-computation}
    \end{center}
    \vspace*{-0.80cm}
\end{table}

\noindent \textbf{Computational overhead}
The computational overhead is determined by the number of integer dilated kernels and the number of convolutional layers whose kernel sizes are not $1\times1$.
As shown in \cref{tab:comp-computation}, the BottleNet in ResNet50~\cite{he-resnet} contains two $1\times1$ kernels and one $3\times3$ kernel.
Since only $3\times3$ kernel introduces the computational overhead, the computational cost for 2 integer dilations is roughly $1.5$ times of the original model, while VVG16~\cite{simonyan-vgg} consists of the majority of $3\times3$ kernels and thus the computation cost is approximately $2$ times.
Some additional computational overhead is caused by the interpolation operation of different dilation kernels. 

\noindent \textbf{Reducing overhead with a grouping strategy}
In practice, the aspect ratios, $\frac{w}{h}$,  of most of images would fall into $[\frac{1}{2}, 2]$, \eg 97.8\% of the training and testing images in the AVA~\cite{murray-ava} dataset.
Training efficiency can be optimized by grouping batches, \eg training with three dilation kernels for the most batches, $DilationRates=\{(2,1), (1,1), (1,2)\}$ for the images whose aspect ratios fall into $[\frac{1}{2}, 2]$.
For the datasets with more diverse aspect ratios, a more fine-grained grouping strategy could be applied. As illustrated in \cref{fig:grouping-strategy}, images with aspect ratio range $[4, 3]$ (above) and $[\frac{1}{2}, 1]$ (below) share the valid integer dilated Convs in the grouped batches.

\noindent \textbf{Parallel optimization}
The calculation of multiple integer dilated kernels in each convolutional layer is equivalent to broadening the output channel size by the number of dilation kernels.
In another words, the computation of dilated Conv group, $\{\kernel{rate_k^i}{rate_k^j}{n} \conv \featurev{n}\}$, can be optimized through parallel computing.
WideResNet~\cite{zagoruyko-wideresnet} claims that increasing the width of Conv layers is more accommodating to the nature of GPU computation and helps effectively balance computations more optimally.
However, from \cref{tab:comp-computation}, the actual training and testing speeds are approximately linearly correlated with \# Muti-Adds, which could be attributed to the current implementation of the framework (TensorFlow) and can be improved by further parallel optimization.

We note that many base networks are stacked mainly with the permutation of $1\times1$ and $3\times3$ kernels and they can be applicable to embed AFDC in terms of the training and inference speed, \ie \cite{he-resnet,huang-densenet,hu-senet,zagoruyko-wideresnet,xie-resnext} in ResNet stream and \cite{andrew-mobilenet,sandler-mobilenetv2,zhang-shufflenet} in MobileNet stream.
Besides, the adaptation is easy because our method is parameter-free.
Overall, the random-size warping preserves the composition of the original image and also provides data augmentation to train the network with scale invariance.
AFDC can adaptively construct fractional dilated kernels according to the spatial distortion information in a computation-efficient manner.

\begin{table*}
    \begin{center}
        \begin{tabular}{p{.5\linewidth}|c|c|c|c|c}
            \hline
            network & cls. acc. & MSE & EMD & SRCC & LCC \\
            \specialrule{1.2pt}{1pt}{1pt}
            NIMA(VGG16)\cite{talebi-nima} & 0.8060 & - & 0.052 & 0.592 & 0.610 \\ 
            NIMA(Inception-v2)\cite{talebi-nima} & 0.8151 & - & 0.050 & 0.612 & 0.636 \\ 
            NIMA(ResNet50, our implementation)& 0.8164 & 0.3169 & 0.0492 & 0.6166 & 0.6388 \\ \hline 
            Vanilla Conv (ResNet50) & 0.8172 & 0.3101 & 0.0481 & 0.6002 & 0.6234 \\ 
            AFDC (random-size cropping pretrain) & 0.8145 & 0.3212 & 0.0520 & 0.6134 & 0.6354 \\ 
            AFDC (aspect-ratio-preserving pretrain) & 0.8295 & 0.2743 & 0.0445 & 0.6410 & 0.6653 \\
            AFDC + SPP & \textbf{0.8324} & \textbf{0.2706} & \textbf{0.0447} & \textbf{0.6489} & \textbf{0.6711} \\ \hline
        \end{tabular} 
        \vspace*{-0.15cm}
        \caption{ \bf Test result comparison on AVA~\cite{murray-ava}: The evaluation metrics are following \cite{talebi-nima}. Reported accuracy values(cls. acc.) are based on binary image classification. MSE(mean squared error), LCC (linear correlation coefficient) and SRCC (Spearman’s rank correlation coefficient) are computed between predicted and ground truth mean scores. EMD measures the closeness of the predicted and ground truth rating distributions with $r=1$ in \cref{eq:emd-loss}. AFDC (random-size cropping) transfers the model trained with widely used data augmentation method in ImageNet, while AFDC (aspect-ratio-preserving pretrain) transfers the model trained with aspect-ratio-preserving data augmentation.}\label{tab:comp-ablation}
    \end{center}
    \vspace*{-0.80cm}
\end{table*}

\subsection{Composition-Aware Structure and Loss}\label{sec:other-modulars}

The commonly-used network structures for the task of image classification usually incorporate global pooling before the fully connected layers~\cite{xie-resnext,he-resnet,huang-densenet,szegedy-googlenet,hu-senet}.
The global pooling eliminates spatial variance which is helpful for the task of image recognition by training the networks with spatial invariant ability, but it causes the loss of localization information for image aesthetics assessment.
Motivated by spatial pyramid pooling~\cite{he-spp}, MNA-CNN-Scene~\cite{mai-adaptive},  several efforts are made to learn the information of spatial image compositions.
First, we use multiple adaptive pooling modules~\cite{he-spp} to output $g_{i}*g_{i}$ grids and feed them into the fully-connected layers (\cf \cref{fig:overview}).
The localization factors for image aesthetics assessment are highly correlated with the image symmetry and the overall image structure.
Then, we aggregate the outputs after the fully-connected layers by concatenation.
To limit the number of model parameters and prevent from overfitting, the module of each adaptive pooling layer outputs $\frac{num_{features}}{num_{grids}}$ channels.

Following the work in \cite{talebi-nima}, we train our network to predict 10-scale score distribution with a softmax function on the top of the network. To get both the mean score prediction and the binary classification prediction, we calculate the weighted sum of score distribution $\sum_{i=1}^{10}i\cdot p_{i}$. We use the ordered distribution distance, Earth Mover Distance~\cite{talebi-nima}, as our loss function:
\begin{equation}\label{eq:emd-loss}
\resizebox{.85\linewidth}{!}
{$
    EMD(p, \hat{p}) = (\frac{1}{N}\sum_{k=1}^{N}|CDF_{p}(k)-CDF_{\hat{p}}(k)|^{r})^{1/r}
$}
\end{equation}
where $CDF_{p}(k)$ is the cumulative distribution function as $\sum_{i=1}^{k}p_{i}$.
As stated in \cref{sec:introduction} and the results in \cite{talebi-nima}, predicting the score distribution can provide more information about image aesthetics compared to the mean scores or binary classification labels.

\section{Experimental Results}
Following \cite{talebi-nima,murray-huberloss,ma-multipath,mai-adaptive,kao2016hierarchical}, we have evaluated our proposed method over AVA dataset~\cite{murray-ava}.
The AVA contains around 250,000 images and each image contains the 10-scale score distribution rated by roughly 200 people.
For a fair comparison, we use the same random split strategy in \cite{talebi-nima,schwarz-tripleloss,murray-huberloss,ma-multipath,mai-adaptive,murray-ava} to generate 235,528 images for training and 20,000 images for test.

\subsection{Implementation Details}\label{subsec:implementation-details}

We use ResNet-50~\cite{he-resnet} as the backbone network due to its efficiency on computation and graphic memory as discussed in \cref{sec:mini-batch}.
We replace all the $3\times3$ Conv layers in each BottleNet with our proposed adaptive fraction dilation Conv layers.
It is easy to plug AFDC into the common CNN architectures since it does not introduce any extra model parameters.
We use the same EMD loss in \cref{eq:emd-loss} with $r=2$ for better back propagation.
To accelerate training, we use the grouping strategy discussed in \cref{sec:mini-batch}.
For the first 12 epochs, we train the model with three dilation kernels, $1\times2$, $1\times1$, $2\times1$ on the grouped images since the aspect ratios for 97.8\% training and validation images fall between $[\frac{1}{2}, 2]$.
Then we train the model with seven dilation kernels, $1\times4$, $1\times3$, $1\times2$, $1\times1$, $2\times1$, $3\times1$, $4\times1$, for the remaining 6 epochs and select the best model from the results in the validation dataset.
We note that the training and test speed could be further accelerated by a more fine-grained grouping strategy.
We transfer the network parameters (pre-trained on ImageNet) before the fully connected layer and set the initial learning rate to $0.01$ for the first 6 epochs.
Then we dampen the learning rate to $0.001$ for the rest of the training epochs.
We find that setting initial learning rate to $0.001$ with a decay rate $0.95$ after every 10 epochs can produce comparable results but converges more slowly.
The weight and bias momentums are set to $0.9$.

\subsection{Ablation Study}
In this section, we introduce the steps to build the final model and analyze the effects of each module step by step: (1) Replacing random cropping with composition-preserving random warping; (2) Replacing vanilla Conv with AFDC in the aspect-ratio-preserving pre-trained model on ImageNet; (3) Adding SPP modules to learn image composition.

\noindent{\bf Random Warping.}
For the data augmentation,  input images in NIMA~\cite{talebi-nima} are rescaled to $256 \times 256$, and then a crop of size $224 \times 224$ is randomly extracted.
They also report that training with random crops without rescaling produces the results that are not compelling due to the inevitable changes in image compositions.
In order to preserve the complete composition, we replace the random-cropping with random-size warping by randomly warping each batch into square size in $[224, 320]$ during each iteration.
The network suffers from overfitting without using random warping.
We note that non-square-size warping may further help with generalization and potentially train AFDC more robustly.

From \cref{tab:comp-ablation}, we generate slightly better results (Vanilla Conv (ResNet50)) compared with NIMA~\cite{talebi-nima}.
We use the same loss (EMD loss) and network (ResNet50, our implementation) as NIMA~\cite{talebi-nima}.
Comparable results have shown that random warping is an effective data augmentation alternative and it preserves the image composition.

\noindent{\bf Aspect-Ratio-Preserving Pretrain.}
We replace the vanilla convolution layers with AFDC in ResNet50.
In our experiments, we find that, fine-tuning the fractional dilated convolution network results in similar validation accuracy compared to the original network (\cf AFDC (random-size cropping pretrain) in \cref{tab:comp-ablation}).
Compatible validation results might be attributed to the pre-trained model which has a distortion-invariant ability.
The widely used data augmentation~\cite{szegedy-googlenet} for network training on ImageNet contains random cropping on a window whose size is distributed evenly between 8\% to 100\% of the original image area with the aspect ratio constrained to $[\frac{3}{4}, \frac{4}{3}]$.
The model is trained with distortion invariance, which has the opposite interest of our method that tries to preserve the original aspect ratio.

For better transfer learning, we pre-train the ResNet50~\cite{he-resnet} on ImageNet~\cite{deng-imagenet} without distortion augmentation.
Specifically, we sample the 8\% to 100\% crop size to the image area with a square window, which is slightly modified comparing to the data augmentation method in~\cite{szegedy-googlenet}.
As in \cref{tab:comp-ablation}, transferring the model from the aspect-ratio-preserving pre-train, we improve the overall test results (AFDC (aspect-ratio-preserving pre-train)) by a margin from the vanilla Conv counterpart.

\begin{figure}
    \begin{center}
        \centering
        \includegraphics[width=.4\linewidth]{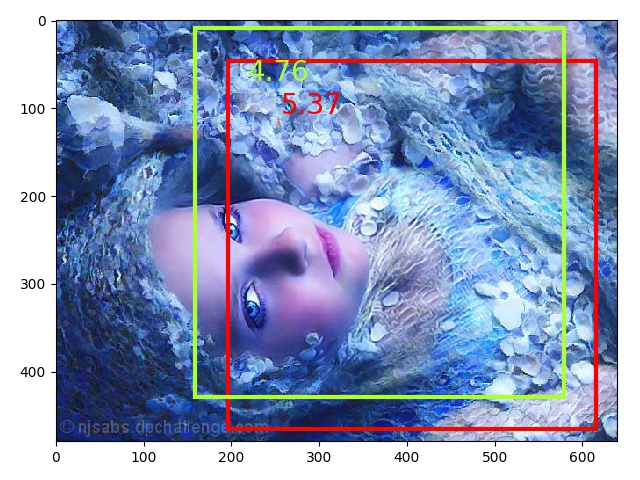}
        \includegraphics[width=.4\linewidth]{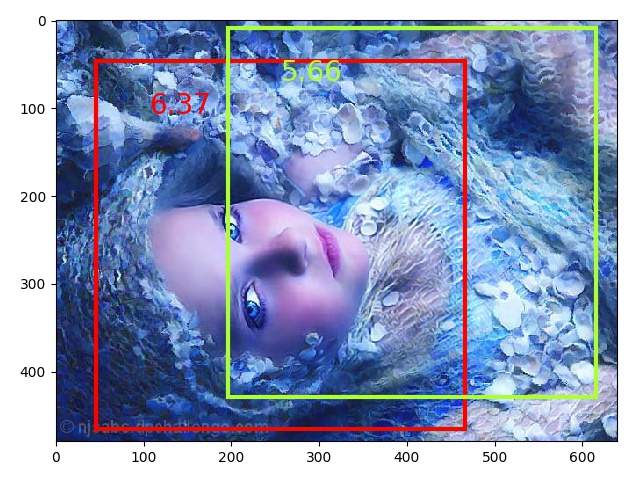}
        \includegraphics[width=.2\linewidth]{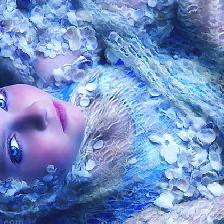}
        \includegraphics[width=.2\linewidth]{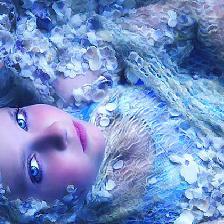}
        \includegraphics[width=.2\linewidth]{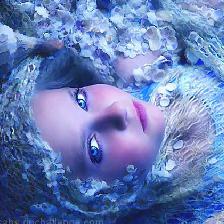}
        \includegraphics[width=.2\linewidth]{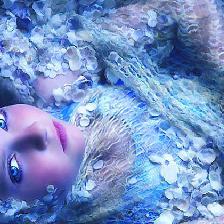}
        
    \end{center} 
    \vspace*{-0.35cm}
    \caption{\bf The cropping results for the model trained with global pooling (left) and SPP (right). The two cropping samples are obtained by using a sliding window with the lowest score (green) and the highest score (red). The image is firstly resized to 256. A sliding window search with size 224 and stride 10 is applied.}
    \label{fig:global-pool-spp} 
    \vspace*{-0.35cm}
\end{figure}

\begin{figure}
    \centering
    \includegraphics[width=.7\linewidth]{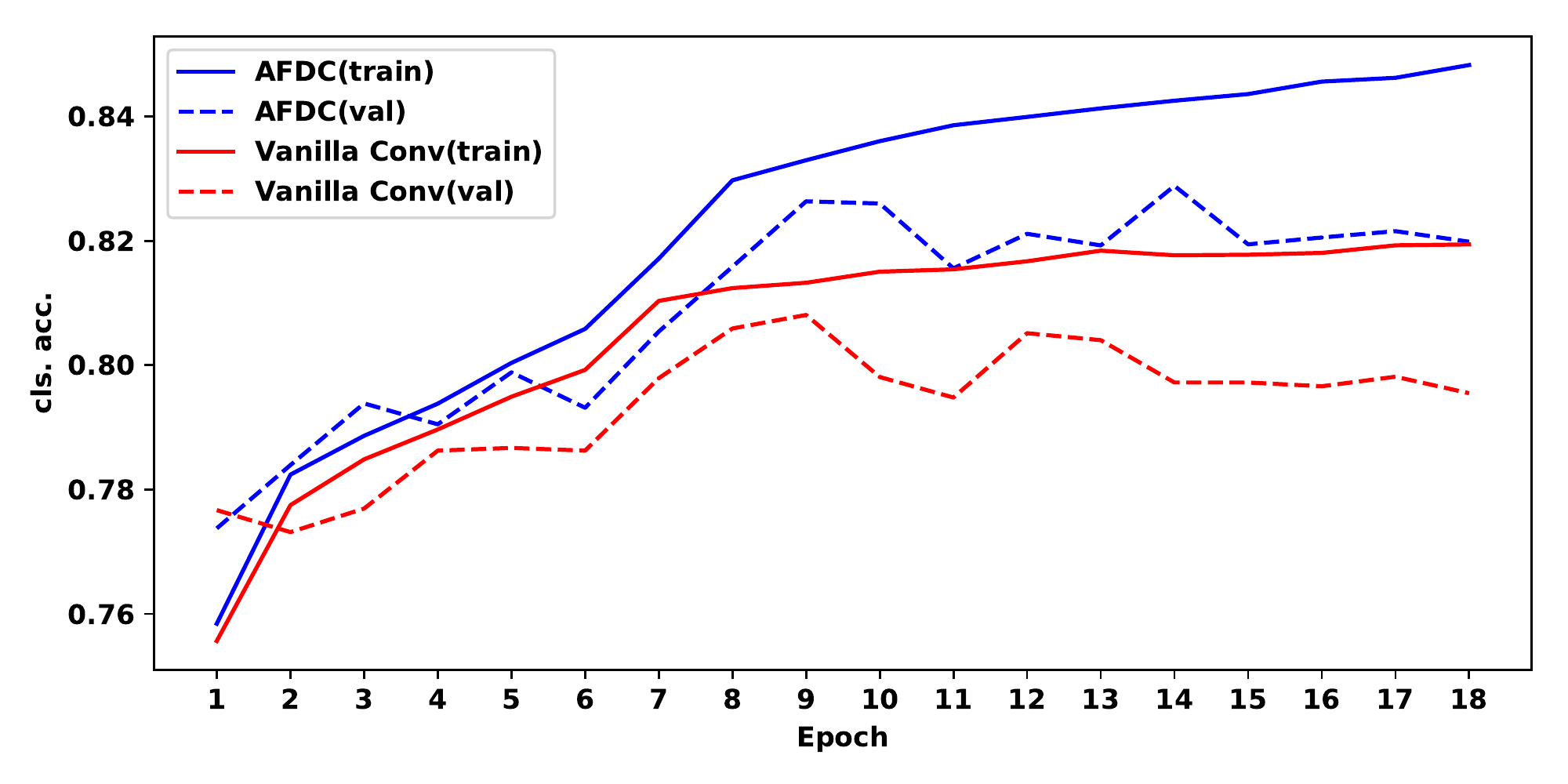}
    \vspace*{-0.35cm}
    \caption{\bf The comparison of learning curves: the backbone networks here are all ResNet-50~\cite{he-resnet}.}
    \label{fig:comp-learning-curves}
    \vspace*{-0.65cm}
\end{figure}

\noindent{\bf Composition-Aware Structure.}
For better representation learning of composition, we use three different scales for SPP, $\{1\times1, 2\times2, 3\times3\}$.
The network with a global pooling layer is equivalent to using only one scale, $1\times1$.
From \cref{tab:comp-ablation}, the network with SPP modules (AFDC+SPP) generates better results comparing to the network with the global pooling layer (AFDC).
The experimental results have shown that incorporating the localization information could benefit the learning of image compositions.
In \cref{fig:global-pool-spp}, the automatic cropping example demonstrates that the ability of localization/composition discrimination is important to find a good cropping result when the global cue in each cropping box has a similar distribution (color, lighting \etal).
The model leaned with SPP modules can infer cropping respecting the image compositions, \eg the relative position of eye and face in the example.
We also tried $num_{grids}=5$ and found that the results were not compelling due to the overfitting from extra model parameters.
Three different scales are quite consistent with the common aesthetic rules (global information, symmetrical composition in horizontal and vertical direction, the rules of the thirds).

\subsection{Effectiveness of AFDC}

\noindent{\bf Learning Representation and Generalization}
    From the experiments in \cref{fig:comp-learning-curves}, we argue that preserving aspect ratio information is essential for learning photo aesthetics since our method not only improves the validation results but also improves the training results.
    Without extra learning parameters, AFDC improves both learning representation and generalization ability.
    As discussed in \cref{sec:introduction}, preserving the image aesthetics information completely omits the label noises caused by random warping and thus facilitates the learning process.
    The additional aesthetic features related to the aspect ratios allow the model to be more robust and discriminative.
    To further probe the effects of embedding aspect ratio, we compare different ways to incorporate the dilated convolution and the results are reported in \cref{tab:comp-dilated-conv}.
    When trained with vanilla Conv (top rows in \cref{tab:comp-dilated-conv}), AFDC is superior to other dilated Conv methods during the test.
    It implies the potential optimal between nearest two integer dilated kernels.
    After training with AFDC (bottom rows in \cref{tab:comp-dilated-conv}), it further validates the effectiveness of AFDC, which is guided by the helpful supervision of aspect ratios.
    We note that such experiments are accessible because our method is parameter-free.
    
    Overall, our proposed AFDC can learn more discriminative and accurate representations related to aesthetics perception, resulting in better generalization by leveraging extra supervision from the information of image aspect ratios.
    
\begin{figure}
    \begin{center}
        \centering
        \includegraphics[width=.9\linewidth]{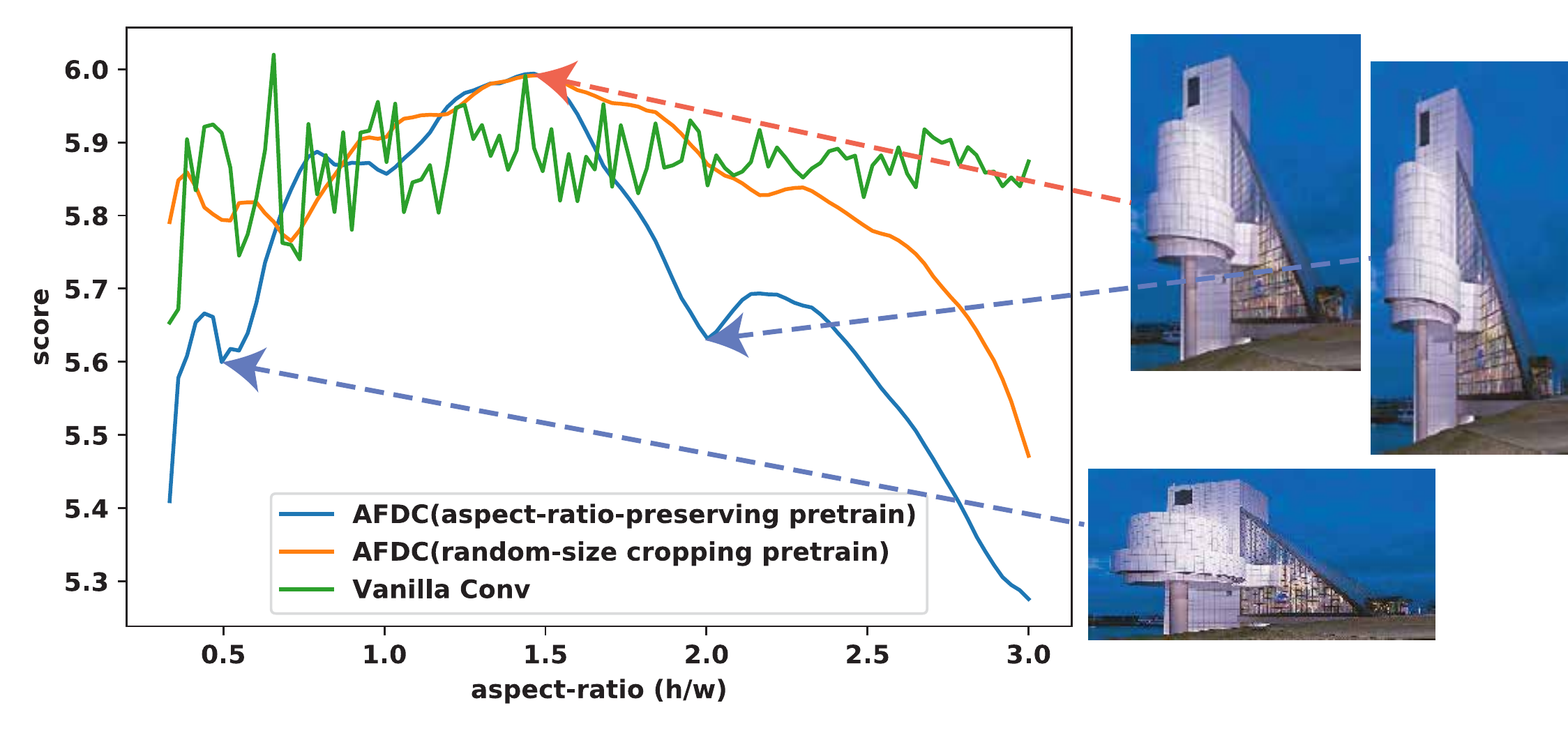}
    \end{center} 
    \vspace*{-0.45cm}
    \caption{\bf Comparison of discrimination to the change of aspect ratios.}
    \label{fig:resize-search} 
    \vspace*{-0.50cm}
\end{figure}

\noindent{\bf Discriminative to Aspect Ratios}
    To further investigate the response to aspect ratios, we resize the same image into different aspect ratios and test the results on different trained models.
    As shown in \cref{fig:resize-search}, AFDC (blue line) is discriminative to the change of aspect ratios.
    The small fluctuation of vanilla Conv (green line) is attributed to sampling change from resizing process.
    The model with random-size cropping pretrain on Imagenet (orange line) is less discriminative to capture the aesthetics perception related to aspect ratio due to its distortion-invariant pretrain.
    Moreover, the proposed method produces a multi-modal score distribution, which reflects that it learns complex relation between the aspect ratio and the aesthetics perception.
    It is in line with the notion that designing better aspect ratios or finding aesthetically pleasing photography angles is not trivial.
    
    Due to the constraint of training dataset, we admit that the learned perception related to the aspect ratios is not satisfactory yet even the model learns from different aspect ratios.
    As a matter of factor, the learning ability is available for our proposed method when training on a more specific targeted dataset.
    It could be utilized in automatic/auxiliary photo enhancement with not only color space transformation but also with spatial transformation, \eg profile editing, multi-shot selection and automatic resizing.

\begin{table}
    \centering
    \scalebox{0.7}{
        \begin{tabular}{c|p{6cm}|cccc}
            \toprule
            Train & Test & cls.acc. & MSE & EMD  \\ 
            \specialrule{1.2pt}{1pt}{1pt} 
            \multirow{6}{*}{vanilla} & vanilla & 0.8172 & 0.3101 & 0.0481  \\
            & constant dilation rate = [2,1] & 0.8072 & 0.5163 & 0.0610   \\
            & second nearest integer dilation & 0.8091 & 0.5368 & 0.0620  \\
            & mean of nearest two integer dilations & 0.8117 & 0.4558 & 0.0576  \\
            & nearest integer dilation  & 0.8114 & 0.4322 & 0.0562  \\
            & adaptive fractional dilation & 0.8132 & 0.4133 & 0.0553  \\ \hline
            \multirow{6}{*}{AFDC} & vanilla & 0.8085 & 0.3210 & 0.0581  \\
            & constant dilation rate = [2,1] & 0.8132 & 0.3182 & 0.0576  \\
            & second nearest integer dilation & 0.8156 & 0.3003 & 0.0476  \\
            & mean of nearest two integer dilations & 0.8274 & 0.2771 & 0.0457  \\
            & nearest integer dilation  &  0.8277 & 0.2757 & 0.0457 \\
            & adaptive fractional dilation & \textbf{0.8295} & \textbf{0.2743} & \textbf{0.0445}  \\
            \bottomrule
        \end{tabular}
    }
    \vspace*{-0.25cm}
    \caption{\bf The test result comparison of different convolutions: The results are obtained with trained parameters by vanilla Conv (above) and AFDC (below). Test processes are conducted by different calculation methods for interpolation weights, $\mathbf{w}$ in \cref{eq:weight-vector}. Vanilla Conv, constant dilation, nearest integer dilation and second nearest integer dilation can be interpreted as feeding one-hot interpolation weight vector into the networks.}
    \label{tab:comp-dilated-conv}
    \vspace*{-0.20cm}
\end{table}

\subsection{Comparison With the State-of-the-Art Results}

\begin{table}
    \begin{center}
        \scalebox{0.85}{
        \begin{tabular}{l|ccc}
            \hline
            Method & cls. acc. & MSE & SRCC \\ 
            \specialrule{1.2pt}{1pt}{1pt} 
            MNA-CNN-Scene~\cite{mai-adaptive} & 76.5\% & - & - \\
            Kong \etal~\cite{kong-ranking} & 77.3\% & - & 0.558 \\
            AMP~\cite{murray-huberloss} & 80.3\% & 0.279 & 0.709  \\
            Zeng \etal (resnet101)~\cite{zeng-unified} & 80.8\% & 0.275 & 0.719 \\
            NIMA (Inception-v2)~\cite{talebi-nima} & 81.5\% & - & 0.612 \\ 
            MP-Net~\cite{ma-multipath} (50 cropping patches) & 81.7\%  & - &- \\ 
            Hosu \etal~\cite{hosu-multi} (20 cropping patches) & 81.7\% & - & \textbf{0.756} \\
            A-Lamp~\cite{ma-multipath} (50 cropping patches) & 82.5\% & - & -\\
            $MP_{ada}$~\cite{sheng-attention}($\ge32$ cropping patches) & 83.0\% & - & - \\ \hline
            ours (single warping patch) & 82.98\% & 0.273 & 0.648  \\
            ours (4 warping patches) & \textbf{83.24\%} & \textbf{0.271} & 0.649 \\
            \hline
        \end{tabular}
        } 
        \vspace*{-0.25cm}
        \caption{\bf Comparison with the SOTA methods: The four patches are warping size $\{224, 256, 288, 320\}$.The single patch is warping size 320 selected from the best results.}
        \label{tab:comp-state-of-the-art}
    \end{center}
    \vspace*{-0.70cm}
\end{table}

We have compared our adaptive fractional dilated CNN with the state-of-the-art methods in \cref{tab:comp-state-of-the-art}.
The results of these methods are directly obtained from the corresponding papers.
As shown in \cref{tab:comp-state-of-the-art}, our proposed AFDC outperforms other methods in terms of cls.acc and MSE, which are the most widely targeted metrics.
Compared with NIMA(Inception-v2)~\cite{talebi-nima} which uses the same EMD loss, our experimental results have shown that preserving the image aesthetic information completely results in better performance on image aesthetics assessment.
We follow the same motivation from MNA-CMM-Scene~\cite{mai-adaptive}, while our proposed method is applicable to mini-batch training which contains images with different aspect ratios.
The experimental results have shown adaptive embedding at kernel level is an effective way to learn more accurate aesthetics perception.
Compared with multi-patch based methods~\cite{ma-multipath,hosu-multi,sheng-attention}, our unified model, which learns the image aesthetic features directly from the complete images in an end-to-end manner, can better preserve the original aesthetic information and alleviate the efforts to aggregate sampling prediction, \eg complicated path sampling strategy and manually designed aggregation structure in \cite{ma-multipath}.
Moreover, our method is much more efficient without feeding multiple cropping patches sampled from original images and could be more applicable for the application.  
Furthermore, it is much succinct due to its parameter-free manner and can be easily adapted to popular CNN architectures.

\section{Conclusion}
In this paper, an adaptive dilated convolution network is developed to explicitly model aspect ratios for image aesthetics assessment.
Our proposed method does not introduce extra model parameters and can be plugged into popular CNN architectures.
Besides, a grouping strategy has been introduced to reduce computational overhead.
Our experimental results have demonstrated the effectiveness of our proposed approach.
Even our adaptive dilated convolution network was proposed to support image aesthetics assessment, it can also be applied in other scenarios when image cropping or warping may introduce label noises.
Moreover, adaptive kernel construction in a parameter-free manner provides an intuitive approach to design dynamic embedding at kernel level, which aims at better learning representation and generalization.

\section*{Acknowledgment}
We would like to thank the anonymous reviewers for their helpful comments.
This work was supported in part by NSFC under Grant (No. 61906143 and No.61473091).

{\small
    \bibliographystyle{ieee_fullname}
    \bibliography{iqabib-new}
}

\end{document}